\documentclass[a4paper]{article}

\usepackage{INTERSPEECH2022}
\usepackage{cite}

\title{Analysis of Self-Attention Head Diversity\\for Conformer-based Automatic Speech Recognition}
\name{Kartik Audhkhasi, Yinghui Huang, Bhuvana Ramabhadran, Pedro J. Moreno}
\address{
  Google LLC, New York}
\email{\{kaudhkhasi, huangyinghui, bhuv, pedro\}@google.com}

\begin{document}
\maketitle
\begin{abstract}
 Attention layers are an integral part of modern end-to-end automatic speech recognition systems, for instance as part of the Transformer or Conformer architecture.
 Attention is typically multi-headed, where each head has an independent set of learned parameters and operates on the same input feature sequence.
 The output of multi-headed attention is a fusion of the outputs from the individual heads.
 We empirically analyze the diversity between representations produced by the different attention heads and demonstrate that the heads become highly correlated during the course of training.
 We investigate a few approaches to increasing attention head diversity, including using different attention mechanisms for each head and auxiliary training loss functions to promote head diversity.
 We show that introducing diversity-promoting auxiliary loss functions during training is a more effective approach, and obtain WER improvements of up to 6\% relative on the Librispeech corpus.
 Finally, we draw a connection between the diversity of attention heads and the similarity of the gradients of head parameters.
\end{abstract}

\section{Introduction}
Attention~\cite{bahdanau2014neural} has become an all-pervasive technology in modern neural network models used in several areas, such as natural language processing and machine translation ~\cite{cho2014learning,young2018recent,gatt2018survey,bahdanau2014neural}, computer vision~\cite{mnih2014recurrent,wang2017residual,xu2015show,gregor2015draw,zhang2019self}, and automatic speech recognition (ASR)~\cite{chorowski2015attention,chan2016listen,sperber2018self,zhang2020transformer,yeh2019transformer}.
A sequence-to-sequence learning problem entails training a model to take a data sequence as input and predict a data sequence as the output. 
A typical multi-layer neural network to solve such a problem predicts a sequence of output representations given a sequence of input representations at each layer.
For each output time step, an \emph{attention mechanism} in a layer assigns a probability density function over the input time sequence, where the probability of any input time step denotes the weight (attention) assigned to it while producing the output representation.

Self-attention~\cite{vaswani2017attention} is the most popular form of attention, and uses the input sequence itself to derive the attention mechanism.
We review dot product self-attention in the next section.
Transformers~\cite{vaswani2017attention} replaced recurrent and convolutional neural networks with self-attention and feed-forward layers, and offered significant speed-ups in training and inference time while providing quality improvements.
It is common practice to use multiple attention mechanisms (\emph{heads})~\cite{lin2017structured} in parallel in each layer and then combine their outputs.
Each attention head uses an independent set of learned parameters.
The intuition behind using multiple heads is that each head is free to attend to different regions of the input sequence and hence provides diverse representations. 
Vaswani et al.~\cite{vaswani2017attention} show a text example where each attention head attends to a different input dependency for the output word (``making'').
Some prior work has investigated the redundancy of attention heads for NLP tasks.
Voita et al.~\cite{voita2019analyzing} show that in a Transformer model, all except a few key attention heads can be pruned without a significant drop in performance on a English-Russian machine translation task.
Kovaleva et al.~\cite{kovaleva2019revealing} make a similar conclusion about BERT attention heads on the GLUE tasks.

A few works in the NLP (especially machine translation) literature have attempted to explicitly promote attention heads to be diverse during model training.
Lin et al.~\cite{lin2017structured} proposed adding a L2 penalty to the training loss that promotes the attention probabilities from different heads to be orthogonal.
They evaluated the models on 3 different NLP tasks -- author profiling, sentiment classification and textual entailment.
Li et al.~\cite{li2018multi} proposed similar loss terms and showed improvements on English-German and Chinese-English machine translation tasks.
Chien et al.~\cite{chien2021disentangled} presented a disentangled masked attention that represents each attention head's probability distribution by a latent topic model.
They additionally minimized an upper-bound on the mutual information between query vectors from pairs of attention heads to encourage them to be diverse.
Correia et al.~\cite{correia2019adaptively} proposed a transformer model with sparse attention heads and adaptive sparsity.
Their machine translation experiments showed that the resulting attention heads have better diversity than the baseline, and also obtain performance improvements.
An et al.~\cite{an2020repulsive} gave a Bayesian view of multi-headed attention and used particle filtering for model training.

To the best of our knowledge, prior work has not analyzed and explicitly improved the diversity of attention heads for the state-of-the-art Conformer~\cite{gulati2020Conformer} (convolution-augmented Transformer) model on automatic speech recognition (ASR).
One related work is by Lohrenz et al.~\cite{lohrenz2021relaxed}, who propose smoothing the attention probability distribution for all heads by interpolating with a uniform distribution, and obtain WER improvements on Librispeech.
We make the following contributions:
\begin{enumerate}
    \item We present an in-depth empirical analysis of the diversity of Conformer attention heads on the public Librispeech corpora and show that multiple attention heads indeed become significantly correlated during training.
    \item We experiment with a few approaches to increasing attention head diversity such as using different attention mechanisms for each head and introducing diversity-promoting loss functions during training. We shows that the latter gives up to 6\% relative improvement in WER.
    \item We analyze the connection between attention head diversity and similarity between gradients of head parameters, and conclude that more diverse heads have less correlated gradients.
\end{enumerate}
The next section gives an overview of multi-headed self-attention.
Section~\ref{sec:head_diversity} presents various diversity losses we used for measuring and promoting attention head diversity during model training.
We presents our experiments in Section~\ref{sec:expts} and conclude the paper in Section~\ref{sec:concl}.

\section{Multi-Headed Self-Attention}\label{sec:mha_review}
We first describe the standard dot product multi-headed self-attention to set the notation and for completeness.
Let $\mathbf{X}$ be the $T \times D$ matrix of $D$-dimensional input vectors over $T$ time steps.
Let $N$ denote the number of attention heads.
The $n^\text{th}$ head computes the following three matrices:

\begin{align}
    \mathbf{Q}_n &= \mathbf{X} \mathbf{W}_n^\text{query} \quad\quad\quad\quad \text{(Query)} \nonumber \\
    \mathbf{K}_n &= \mathbf{X} \mathbf{W}_n^\text{key} \quad\quad\quad\quad\quad \text{(Key)} \nonumber \\
    \mathbf{V}_n &= \mathbf{X} \mathbf{W}_n^\text{value} \quad\quad\quad\quad \text{(Value)}
\end{align}
where $\mathbf{W}_n^\text{query}$, $\mathbf{W}_n^\text{key}$ and $\mathbf{W}_n^\text{value}$ are $D \times H$ weight matrices that transform the input matrix $\mathbf{X}$ into three $T \times H$ matrices -- $\mathbf{Q}_n$ (query), $\mathbf{K}_n$ (key), and $\mathbf{V}_n$ (value).
We then compute dot product between the query and key matrices followed by row-wise softmax to obtain the $T \times T$ attention probability matrix
\begin{align}
    \mathbf{A}_n &= \text{softmax}(\mathbf{Q}_n \mathbf{K}_n^T / H) \;.
\end{align}
The value matrix is then multiplied by the attention probability matrix to obtain the $T \times H$ output \emph{context} matrix from the $n^\text{th}$ head:
\begin{align}
    \mathbf{Y}_n &= \mathbf{A}_n \mathbf{V}_n \;.
\end{align}
The $t^\text{th}$ row of $\mathbf{Y}_n$ is a convex combination of all rows of $\mathbf{V}_n$ using attention weights given by the $t^\text{th}$ row of $\mathbf{A}$.
Next, the output matrices from the $N$ attention heads are concatenated along rows and down-projected to produce the final $T \times D$ output of the attention layer:
\begin{align}
    \mathbf{Y} &= [\mathbf{Y}_1\quad\mathbf{Y}_2\ldots\mathbf{Y}_N] \mathbf{W}^\text{out} 
\end{align}
where $\mathbf{W}^\text{out}$ is the $NH \times D$ projection matrix.
The next section discusses various losses we used for measuring and promoting self-attention head diversity.

\section{Measuring and Promoting Attention Head Diversity}\label{sec:head_diversity}
For the sake of illustration, consider the computation of diversity between the $T \times H$ output representations (\emph{context vectors}) $\mathbf{Y}_m$ and $\mathbf{Y}_n$ produced by the $m^\text{th}$ and $n^\text{th}$ attention heads.
We compute the correlation coefficient between two heads $m$ and $n$ as
\begin{align}\label{eq:cosine_sim}
    d^\text{Y}(m, n) &= \frac{1}{T}\text{sum}\big{(}\mathbf{\tilde{Y}}_m \odot \mathbf{\tilde{Y}}_n\big{)}
\end{align}
where $\mathbf{\tilde{Y}_m}$ and $\mathbf{\tilde{Y}_n}$ denote the matrices of row unit vectors, $\odot$ denotes the element-wise product of two matrices and $\text{sum}$ is the sum of all entries.
Given an $N \times N$ matrix of the above correlation coefficients, we would like to compute a scalar loss that captures the average diversity of the $N$ heads.
Each $d^\text{Y}(m, n) \in [-1, 1]$ and correlation coefficients of both $1$ and $-1$ would correspond to minimal diversity.
In other words, perfectly diverse heads produce perfectly decorrelated/orthogonal representations.
Hence, similar to~\cite{lin2017structured}, we define the following diversity loss:
\begin{align}\label{eq:div_loss}
    \mathcal{L}^\text{diversity} &= \frac{1}{N^2} \sum_{m=1}^N \sum_{n=1}^N \big{(}d^\text{Y}(m, n) - I(m, n)\big{)}^2
\end{align}
where $I(m, n)$ denotes the $(m, n)^\text{th}$ entry of the identity matrix.
This loss is proportional to the Frobenius norm of the difference between the $N \times N$ diversity loss and identity matrices.
\begin{table}[h]
\centering
\begin{tabular}{ |c|c| } 
 \hline
 \textbf{Representation Name} & \textbf{Diversity loss} \\ \hline
 Context ($\mathbf{Y}$) & $d^\text{Y}(m, n)$ \\ \hline
 Attention Probability ($\mathbf{A}$) & $d^\text{A}(m, n)$ \\ \hline
 Query ($\mathbf{Q}$) & $d^\text{Q}(m, n)$ \\ \hline
 Key ($\mathbf{K}$) & $d^\text{K}(m, n)$ \\ \hline
 Value ($\mathbf{V}$) & $d^\text{V}(m, n)$ \\ \hline
\end{tabular}
\caption{Various attention diversity losses used in this paper.}
\label{tab:div_scores}
\end{table}

Multi-headed attention provides several representations for computing the diversity loss as defined above. Table~\ref{tab:div_scores} lists the diversity losses we used.
In order to promote diversity during training, we add the  diversity loss as an auxiliary loss function:
\begin{align}\label{eq:tot_loss}
    \mathcal{L}^\text{total} &= \mathcal{L}^\text{RNNT} + \lambda \mathcal{L}^\text{diversity} 
\end{align}
where $\lambda$ is a scalar hyper-parameter and $\mathcal{L}^\text{RNNT}$ denotes the Recurrent Neural Network Transducer (RNNT) loss, defined as the negative log-likelihood of the output label sequence given the input acoustic feature sequence. 
We also considered using the negative Kullback-Liebler (KL) divergence as the diversity loss $d^\text{A}(m, n)$ between the attention probability matrices.
However, that would lead to unstable training because we would like to maximize KL divergence and it is unbounded from above.

\section{Experiments}\label{sec:expts}
\subsection{Dataset and Model Architecture}
We use the well-benchmarked, publicly-available Librispeech~\cite{panayotov2015librispeech} corpus for all our experiments. 
Librispeech comprises of read 960 hours of speech from audiobooks from over 2000 speakers.
We did not include any additional audio or text data in our experiments.

The Conformer-RNNT model follows an encoder-decoder architecture detailed in~\cite{gulati2020Conformer} and consists of a 17-layer Conformer acoustic encoder with self-attention in all layers, a 2-layer unidirectional LSTM decoder, and a joint network.
The full network is trained end-to-end using the RNN-T loss.
The Conformer acoustic encoder uses multi-headed attention over the whole utterance with 8 attention heads.
We use a model dimension of 512 for the encoder, which results in a total of 118M parameters. 
The input acoustic features for our LibriSpeech experiments are the same as used in~\cite{gulati2020Conformer}. 
The target vocabulary for all models are wordpieces with a vocabulary of 1024. 
All models are randomly-initialized and trained with an effective batch size of 4096 in Lingvo~\cite{shen2019lingvo} on Tensor Processing Unit slices. 
Our ablation experiments did not use relative positional embedding in the attention layer.
We later show results with one setting on a model with relative positional embedding.

Table~\ref{tab:baseline} shows several baseline models with different attention context sizes.
We next present the analysis of attention head diversity for the full context baseline system.

\begin{table}[h]
\centering
\begin{tabular}{|c|c|c|c|c|} 
 \hline
 \textbf{Attention} & \textbf{dev} & \textbf{dev-other} & \textbf{test} & \textbf{test-other} \\ \hline
 Full context & 2.0 & 5.1 & 2.2 & 5.2 \\ \hline
 L=256, R=256 & 2.1 & 5.0 & 2.2 & 5.2 \\ \hline
 L=128, R=128 & 2.1 & 5.3 & 2.2 & 5.3 \\ \hline
 L=64, R=64 & 2.1 & 5.7 & 2.4 & 5.7 \\ \hline
 L=32, R=32 & 2.2 & 6.0 & 2.4 & 6.0 \\ \hline
 L=16, R=16 & 2.3 & 6.7 & 2.4 & 6.5 \\ \hline
\end{tabular}
\caption{Baseline WERs for different left (L) and right (R) context sizes measured in terms of time steps.}
\label{tab:baseline}
\end{table}

\subsection{Analysis of Attention Head Diversity}
We computed various attention head diversity losses presented in (\ref{eq:div_loss}) and Table~\ref{tab:div_scores} using the baseline model for the Librispeech dev and test sets at the convergence of training.
Note that these scores are summed over all 17 layers of the Conformer acoustic encoder.

We conclude from Table~\ref{tab:div_scores_baseline} that the attention probabilities from the various heads show the highest diversity loss and hence the highest correlation.
This implies that the different attention heads are often focusing on the same frames of the input sequence.
The query and key diversity losses are the next highest, though significantly lower than the attention probability diversity loss.
The value and context vector diversity losses are the lowest.
\begin{table}
\centering
\begin{tabular}{ |c|c|c|c|c| } 
 \hline
 \textbf{Diversity loss} & \textbf{dev} & \textbf{dev-other} & \textbf{test} & \textbf{test-other} \\ \hline
 $d^\text{A}(m, n)$ & 6.37 & 6.02 & 6.31 & 6.10 \\ \hline
 $d^\text{Q}(m, n)$ & 0.53 & 0.59 & 0.54 & 0.55 \\ \hline
 $d^\text{K}(m, n)$ & 0.57 & 0.61 & 0.61 & 0.58 \\ \hline
 $d^\text{V}(m, n)$ & 0.13 & 0.14 & 0.13 & 0.14 \\ \hline
 $d^\text{Y}(m, n)$ & 0.19 & 0.20 & 0.20 & 0.20 \\ \hline
\end{tabular}
\caption{Attention diversity losses summed over all layers of the Conformer acoustic encoder for the baseline full-context Librispeech model.}
\label{tab:div_scores_baseline}
\end{table}
We next discuss a simple method to introduce attention head diversity -- by using a mixture of different attention mechanisms. 

\subsection{Mixture of Different Attention Mechanisms}
Intuitively, one would expect that using different attention mechanisms for subsets of heads would automatically generate enough diversity, and would result in a better WER than the baseline.
In order to check this hypothesis, we trained our models with the following mixtures of different attention mechanisms:
\begin{enumerate}
    \item \textbf{Different left and right contexts} - We mixed some attention mechanisms with different left (L) and right (R) context widths from Table~\ref{tab:baseline}.
    \item \textbf{Multi-headed softmax and FAVOR attention} - We mixed full context multi-headed softmax attention and FAVOR attention, which is used in Performers. FAVOR attention~\cite{choromanski2020rethinking} approximates softmax attention by computing query-key similarity via dot product in a random kernel space. Since FAVOR attention uses a very different way to compute dot products versus standard softmax attention, we hoped that combining the two would naturally lead to some diversity.
\end{enumerate}
Table~\ref{tab:atten_mix} shows the WERs for a few relevant baselines from Table~\ref{tab:baseline} with the same form of attention mechanism used for all 8 heads.
It also shows the WERs for a few attention mixtures.
We observe that none of the attention mechanism mixtures are able to improve the WER over the baseline models.
This indicates that simply combining different attention mechanisms is not enough to ensure diversity of the attention heads.

\begin{table}
\centering
\begin{tabular}{|c|c|c|c|c|} 
 \hline
 \textbf{Attention} & \textbf{dev} & \textbf{dev-other} & \textbf{test} & \textbf{test-other} \\ 
 \textbf{:Number of Heads} & & & & \\ \hline
 \multicolumn{5}{|c|}{\emph{Full context}} \\ \hline
 Full-ctx Softmax: 8 & 2.0 & 5.1 & 2.2 & 5.2 \\ \hline
 Full-ctx Softmax: 4 & 2.0 & 5.1 & 2.2 & 5.1 \\
 FAVOR: 4 & & & & \\ \hline\hline
 \multicolumn{5}{|c|}{\emph{Limited context}} \\ \hline
 (L=256, R=256): 8 & 2.1 & 5.0 & 2.2 & 5.2 \\ \hline
 (L=256, R=256): 4 & 2.1 & 5.3 & 2.3 & 5.4 \\
 (L=256, R=0): 4 & & & & \\ \hline
 (L=128, R=128): 8 & 2.1 & 5.3 & 2.2 & 5.3 \\ \hline
 (L=128, R=128): 2 & 2.1 & \textbf{5.2} & 2.3 & 5.4 \\ 
 (L=64, R=64): 2 & & & & \\ 
 (L=32, R=32): 2 & & & & \\ 
 (L=16, R=16): 2 & & & & \\ \hline
\end{tabular}
\caption{WERs for mixtures of different attention mechanisms. The rows with all 8 attention heads using the same attention mechanism are the baselines.}
\label{tab:atten_mix}
\end{table}

\subsection{Diversity-promoting Auxiliary Training Losses}
Next, we explored the impact of explicitly introducing diversity promoting auxiliary loss functions during model training.
We focused on the full context softmax Conformer model in this section.
We experimented with each diversity loss in Table~\ref{tab:div_scores} as an auxiliary loss function during RNNT training as shown in (\ref{eq:tot_loss}).
We tuned the weight $\lambda$ of this auxiliary loss on the dev and dev-other test sets.
Table~\ref{tab:div_loss_wer} shows the Librispeech WERs.
We observe that including diversity losses seems to consistently improve the WER over the baseline model.
\begin{table}[h]
\centering
\begin{tabular}{ |c|c|c|c|c| } 
 \hline
 \textbf{Diversity score} & \textbf{dev} & \textbf{dev-other} & \textbf{test} & \textbf{test-other} \\ \hline
 Baseline & 2.0 & 5.1 & 2.2 & 5.2 \\ \hline
 $d^\text{A}(m, n)$ & 2.0 & \textbf{4.8} & \textbf{2.1} & \textbf{5.1} \\ \hline
 $d^\text{Y}(m, n)$ & 2.0 & \textbf{4.8} & 2.2 & \textbf{5.1} \\ \hline
 $d^\text{Q}(m, n)$ & 2.0 & \textbf{4.8} & \textbf{2.1} & \textbf{5.0} \\ \hline
 $d^\text{K}(m, n)$ & 2.1 & \textbf{4.9} & \textbf{2.1} & \textbf{5.1} \\ \hline
 $d^\text{V}(m, n)$ & 2.1 & \textbf{4.9} & \textbf{2.1} & \textbf{5.0} \\ \hline
\end{tabular}
\caption{WERs using different diversity scores as auxiliary loss functions during training. All models use the full context. }
\label{tab:div_loss_wer}
\end{table}

Next, we analyzed the impact of including the diversity auxiliary loss function during training on the diversity losses themselves.
Table~\ref{tab:div_score_after_train} shows the results.
When compared to Table~\ref{tab:div_scores_baseline}, we observe that all the diversity losses are significantly reduced, indicating that the different attention heads are indeed diverse. 
Attention probabilities are the most directly-interpretable representations computed in an attention layer. 
The reduction in its diversity loss from approximately 6.0 (Table~\ref{tab:div_scores_baseline}) to 0.4 (Table~\ref{tab:div_score_after_train}) indicates that the different attention heads are now focusing on different subsets of input frames.
\begin{table}[h]
\centering
\begin{tabular}{ |c|c|c|c|c| } 
 \hline
  \textbf{Diversity loss} & \textbf{dev} & \textbf{dev-other} & \textbf{test} & \textbf{test-other} \\ \hline
 $d^\text{A}(m, n)$ & 0.41 & 0.37 & 0.45 & 0.39 \\ \hline
 $d^\text{Q}(m, n)$ & 0.02 & 0.02 & 0.02 & 0.02 \\ \hline
 $d^\text{K}(m, n)$ & 0.00 & 0.00 & 0.00 & 0.00 \\ \hline
 $d^\text{V}(m, n)$ & 0.01 & 0.01 & 0.01 & 0.01 \\ \hline
 $d^\text{Y}(m, n)$ & 0.04 & 0.05 & 0.05 & 0.05 \\ \hline
\end{tabular}
\caption{Attention diversity scores summed over all layers of the Conformer encoder for the models trained with diversity loss.}
\label{tab:div_score_after_train}
\end{table}
\begin{figure*}[t]
\includegraphics[width=14.5cm]{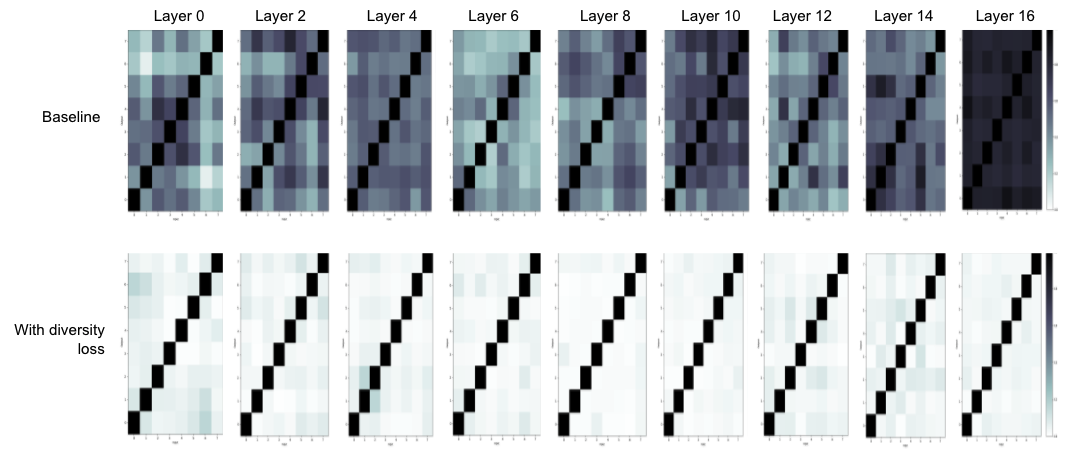}
\centering
\caption{This figure shows the cosine similarity between the 8 attention heads for the baseline model and the model trained with the attention probability diversity loss. We show images only for the even-numbered Conformer layers. Darker colors imply higher cosine similarity (closer to 1).}
\label{fig:head_prob_corr_images}
\end{figure*}

To further understand the impact of training with the attention probability diversity loss, we computed the $N \times N$ cosine similarity matrix of the $N$ attention heads for each Conformer layer on a Librispeech utterance.
Figure~\ref{fig:head_prob_corr_images} shows the $8 \times 8$ cosine similarity matrix for the even-numbered layers of the Conformer model using both the baseline model and the model trained with attention probability diversity loss.
We observer that the baseline model contains highly correlated and redundant attention heads.
In particular the final layer (layer 16) shows an extremely high correlation between all pairs of attention heads.
The attention heads for each layer become significantly more decorrelated upon training with the diversity loss, and this also results in a WER improvement.

Finally, we trained a different and stronger Conformer model with relative position encodings and evaluated the impact of using the attention context diversity loss.
Table~\ref{tab:relpos_wer} shows the WERs of the two models.
We observe that including the diversity auxiliary loss function obtains a WER improvement even over a stronger baseline model on the harder dev-other and test-other sets.
\begin{table}[h]
\centering
\begin{tabular}{ |c|c|c|c|c| } 
 \hline
 \textbf{Diversity score} & \textbf{dev} & \textbf{dev-other} & \textbf{test} & \textbf{test-other} \\ \hline
 Baseline & 1.9 & 4.3 & 2.1 & 4.6 \\ \hline
 $d^\text{Y}(m, n)$ & 1.9 & \textbf{4.2} & 2.1 & \textbf{4.4} \\ \hline
\end{tabular}
\caption{WERs using a baseline Conformer model with relative position encoding and a model using context diversity loss as auxiliary loss functions during training. }
\label{tab:relpos_wer}
\end{table}
\vspace{-15pt}
\subsection{Analysis of gradient similarity}
We next asked the question -- Do more diverse attention heads also receive more diverse gradients during backpropagation for a given batch of data?
We focused on the query weight matrix for the purposes of illustration.
We considered three models with different choices of the diversity loss weight $\lambda$ -- baseline ($\lambda = 0$), $\lambda = 0.001$, and $\lambda = 1.0$. 
Higher values of $\lambda$ force the attention heads to be more diverse during training.

Given a pair $(m, n)$ of attention heads, we computed the gradient of the query weight matrices $\mathbf{W}_m^\text{query}$ and $\mathbf{W}_n^\text{query}$ using a batch of training data.
In similar fashion to the diversity loss computation in (\ref{eq:cosine_sim}) and (\ref{eq:div_loss}), we first computed the cosine similarity between the gradient tensors, and then computed the mean of squares of the $N \times N$ cosine similarity matrix minus the identity matrix over all possible pairs of attention heads and layers.
\begin{table}[h]
\centering
\begin{tabular}{ |c|c|c|c|c| } 
 \hline
 \textbf{$\lambda$} & \textbf{dev} & \textbf{dev-other} & \textbf{test} & \textbf{test-other} \\ \hline
 0 & 0.021 & 0.026 & 0.021 & 0.026 \\ \hline
 0.001 & 0.016 & 0.017 & 0.016 & 0.018 \\ \hline
 1.0 & 0.015 & 0.017 & 0.015 & 0.017 \\ \hline
\end{tabular}
\caption{Gradient similarity loss between query weight matrices for different choices of query diversity loss weight ($\lambda$) used during model training.}
\label{tab:grad_sim}
\end{table}
Table~\ref{tab:grad_sim} shows the query gradient similarity loss for the baseline ($\lambda = 0$) and two models trained with diversity loss.
We observe that the model trained with the most weight ($\lambda = 1.0$) to the diversity loss also saw the most diverse gradients received by the individual heads, as indicated by the lowest value of the gradient diversity loss.
Hence, the weight matrices of more diverse attention heads update very differently given the same batch of data, when compared to less diverse attention heads. 

\section{Conclusion}\label{sec:concl}
We analysed attention head diversity for one of the state-of-the-art models used for ASR -- Conformer, on the public Librispeech data set.
We evaluated diversity losses computed over all intermediate and final representations in each self-attention layer of the model. 
The paper showed that the various attention heads do tend to become highly correlated during the course of model training.
In particular, the attention probabilities show the highest correlation across attention heads.
We then show that adding auxiliary training loss functions that promote head diversity do make the heads less redundant and improve the model's WER.
Finally, we evaluated the connection between attention head diversity and gradient similarity of the head weight matrices.
We showed that more diverse attention heads also received more diverse gradients.

\bibliographystyle{IEEEtran}

\bibliography{mybib}

\end{document}